# $C^3$A: A Cognitive Collaborative Control Architecture For an Intelligent Wheelchair

Rupam Bhattacharyya [a,*], Adity Saikia [a] and Shyamanta M Hazarika [a]

[a] *Department of Computer Science and Engineering, Tezpur University, Tezpur, 784028, India*
*E-mail: rupam15@tezu.ernet.in*

**Abstract.** Retention of residual skills for persons who partially lose their cognitive or physical ability is of utmost importance. Research is focused on developing systems that provide need-based assistance for retention of such residual skills. This paper describes a novel cognitive collaborative control architecture $C^3$A, designed to address the challenges of developing need- based assistance for wheelchair navigation. Organization of $C^3$A is detailed and results from simulation of the proposed architecture is presented. For simulation of our proposed architecture, we have used ROS (Robot Operating System) as a control framework and a 3D robotic simulator called USARSim (Unified System for Automation and Robot Simulation).

## 1. Introduction

Recent progresses in the field of humanitarian robotics and automation technology has brought smiles to millions of people who suffer from cognitive and mobility impairment. In spite of tremendous progress in the area of intelligent assistive devices, intelligent wheelchairs are yet to be widely accepted. According to a survey consisting of 200 clinicians, greater than 50% of wheelchair users reported complaint with its control [23][5]. Further, persons who lose their cognitive/physical ability have to go through a rehearsal of their residual capability in a continuous manner. This prevents them from losing any further skill set and may rejuvenate them to acquire the lost abilities. For this reason, the assistance to human patient must be offered on need basis [15]. We aspire to offer a wheelchair which can be used by people having different levels of cognitive and/or physical impairment.

A novel cognitive collaborative control architecture is presented in this paper. Collaboration is best explained by the following examples. Let us consider that a human patient is currently sitting on an intelligent wheelchair with robotic arms. Their shared goal is to lift the pen placed on the table. Suppose the human patient is active and moves near to the table; suddenly the human is no longer able to proceed further towards the goal. This situation is immediately traced by the robotic wheelchair and lifts

*Corresponding author. E-mail: rupam15@tezu.ernet.in Tel.: +91-3712-275136 Fax: +91-3712-267005

the pen from the table. For other type of collaboration, consider two persons are riding a tandem bicycle. If one person slows down, then other person has to put some extra effort to maintain the balance and to move on to the common destination. In both the examples, the adaptation is automatic because of which collaboration is so delightful and effective. The conceptual understanding of the two terms; namely “shared control" and “collaborative control" is essential for the readers to understand our architecture in a better way. Following conceptual definitions of the two terms provided by Urdiales [8] are considered as the standard one and the remaining section of this paper will follow these only.

**Shared control:**

“Situations where machines and persons cooperate to achieve a common goal fall within the field of shared control.”[8]

**Collaborative control:**

“Collaborative control is a specific type of shared control where there are no sharp control switches between person and machine. Furthermore, humans always retain some control and are rewarded with more when they perform better, yet receive more assistance when needed.”[8]

The remainder of the paper is organized as follows. In section 2, we discuss the human-robot co-operative approaches present in the literature and the requirements needed to design a cognitive agent. Section 3 describes $C^3$A: a three layered cognitive collaborative control architecture. Section 4 is divided into three subsections; namely Evaluation Metrics, Experimental Set Up, and Experimentation. In simulation of our proposed architecture, we have used USARSim [35] and ROS [28]. P3AT robot is treated as our wheelchair which imbibes our architecture. This research work envisages the integration of USARSim/ROS to build a cognitive agent on top of $C^3$A. Section 3 and Section 4 collectively gives an insight to the systematic approach of designing and implementing the whole system. Section 5 deals with performance of the whole system. Concluding remarks and future trend will be discussed in section 6.

## 2. Background

### *2.1. Approaches related to human-robot co-operation*

Co-operation between human and robot addresses several relationships between them. As there is a considerable difference in users’ ability to control a wheelchair, the system’s contribution to control plays a significant role. According to [9], control approaches in which human and robots (or machine) jointly operate with each other to achieve a common destination can be classified into:

- safeguarded operation and
- shared control.

In safeguarded operation, robots will be completely controlled by the user and robot occasionally takes over to avoid possible danger [9]. Table 1 will help the readers to understand the basic principle of a few well known shared control approaches for human-robot co-operation. It is difficult to discuss all these approaches and their differences with collaborative control. Two survey papers namely [24] and [1]; are good starting point to start research in human-robot co-operation/collaboration. From the discussion till now; collaborative control seems to be the best for rehabilitation of patients.

Urdiales et. al. [9] can be considered as pioneer in collaborative control strategy for wheelchair. A number of smart wheelchairs - MAID, NavChair, TinMan and SmartChair use shared control [34]; differing only in how behaviors are implemented. There are two intelligent wheelchair technologies namely Smile Rehab[1] and TAO-7 Intelligent Wheelchair Base from AAI Canada[2] which made the transition from the lab environment to the public and commercial sectors [40]. A common problem in most of these systems is that the man and the machine do not contribute to control simultaneously. Though see [9]; wherein this is overcome using a purely reactive navigation system [32]. Urdiales et. al. [9] report detailed clinical trails of their collaborative system based on reactive navigation. In [39], the collaborative controller uses a multiple-hypotheses method to continuously predicts the short-term goals of the user and calculates an associated confidence of the prediction.

Table 1 : Different shared control approaches of human-robot co-operation

| Approach | Co-operative strategy (intuitive definition from the literature) | Remark |
|---|---|---|
| Supervisory control | It is described as the concept in which control is performed by an intelligent controller under the supervision of a human instead of the human performing direct manual control [22]. | Supervisory control maybe necessary when the human and the robot are geographically separated or when a single human supervises a large number of robots. |
| Adjustable autonomy | Enable the individual robots to act fairly independently of one another, while still allowing for tight, precise coordination when necessary [10]. | Designed for multi-robot system |
| Sliding autonomy | Decisions about when to switch between autonomous and tele-operated control can be made smoothly, both by humans and by the robots themselves.<br><br>The autonomous system is given the ability to ask for human assistance[17] [14]. | Enable multiple heterogeneous robots to work together, and with humans. |
| Symbiotic relationship | The robot accomplishes tasks for humans and ask for help only to finish the task perfectly [38]. | A visitor-companion robot is built using this strategy. |
| Mixed-initiative interaction(MII) | Mixed-initiative interaction lets agents work most effectively as a team and agents dynamically adapt their interaction style to best address the problem at hand [18].<br><br>Various definitions of MII is present in literature. | In [16], a human-robot collaborative architecture is built using the MII approach. |
| Coactive design | Coactive Design is a teamwork-centered approach. The fundamental principle of Coactive Design recognizes that the underlying interdependence of participants in joint activity is a critical factor in the design of human-agent systems[25]. | Collaborative Control is a first step toward Coactive Design. |

It is observed that literature related to the stability analysis of wheelchair controllers based on shared control approaches is scarce. Novel techniques such as [41] and [42] can be a good starting point to undergo such an analysis. With the availability of brain computer interfaces (BCI), a wheelchair can be controlled through EEG signals [43]. Similar problem in a distributed environment has been studied in [44].

### *2.2. Cognitive agent: requirement and design principle*

Embodiment of cognition within an intelligent agent will bolster human robot interaction [2]. To generate rational behaviour in any external environment, agent should be embedded with well established cognitive architecture like ACT-R [19] or SOAR [20]. The conceptual understanding behind a *rational agent* and *its behaviour* in our context is provided below:
**Rational agent:** "A system that makes decisions by considering all possible actions and choosing the one that leads to the best expected outcome can be treated as rational agent."[37]
**Rational behavior:** Rational behavior in human-robot collaboration will be to assist the user whenever required and our cognitive agent help us to attain this behavior.
However the selection of a particular cognitive architecture to tackle a distinct HCI problem is hugely difficult [30]. Our aim is to use a popular cognitive architecture and introduce within it certain novel features to effect collaboration as these architectures don't have any module for collaboration per se. This certainly opens up the path of research in building a cognitive model incorporating collaboration into it. ACT-R has been utilized successfully to model higher-level cognition phenomena, such as memory, reasoning and skill acquisition [21]. This is the main reason that encourage us to incorporate the flavor of ACT-R into the $C^3$A.

To imitate human cognition, cognitive agents can be constructed to act properly in changing environment. Throughout this paper, we have used cognitive agent, wheelchair and P3AT robot interchangeably. The cognitive agent can be defined as below [6]:

- Functions incessantly and autonomously in an environment.
- Able to accomplish activities in an adaptable and intelligent manner.
- Responsive to modifications in the environment.
- Proactive; exhibit goal-oriented and opportunistic behaviour.
- Take the initiative when required.
- Learn from experience.

---

[1] http://www.smilesmart-tech.com/assistive-technology-products/smile-smart-wheelchair/
[2] http://www.aai.ca/robots/tao_7.html

There must be some combination of the agent's data structures and algorithms must reflect the knowledge it has about its surrounding environment [26]. Agent architectures can be divided into three broad categories based on [29]; we are interested in the reactive architecture which is defined to be the one that does not include any kind of central symbolic world model, and does not use complex symbolic reasoning [29]. A subsumption architecture [32] is a hierarchy of task-accomplishing behaviours which followed this alternative approach. Thus a compact collaborative control architecture with the flavor of ACT-R may be helpful in fulfilling a decent acceptance rate among users along with simultaneous control and need-based assistance.

## 3. $C^3A$ : Cognitive Collaborative Control Architecture

The proposed cognitive collaborative control architecture is shown in Figure 1. It is inspired from the work [11] and adaptation of well established cognitive architecture, ACT-R [19]. $C^3A$ is a layered architecture with the following layers:

- User Interface Layer,
- Superior Control Layer and
- Local Control Layer.

The functionality of each layer and the inter dependence among them are described below.

### *3.1. User Interface Layer*

User has to interact with this layer for setting the shared goal in the environment. We have assumed that the user of our system is visually active. User Interface Layer helps the users to navigate the P3AT through the maze environment with the help of keyboard. This layer is utilized by the users for interacting with rest of the layers of $C^3A$. The wheelchair will be used by people who have either physical disability or cognitive disability or both. So, it is essential for us to determine how much help the patient should be provided so that their residual skill remains intact. There are various effective tests to measure the current health status of the patient. In [9], they have conducted five tests to calculate the cognitive/physical disability score. The procedure followed by us to calculate the score can be found in the subsection 4.3. This score is stored in procedural memory of the memory module for future usage. Later on, the score is fed to the heuristic engine of the reactive module for further processing.

### *3.2. Superior Control Layer*

This layer is composed of two sub-modules;

- Reactive Module and
- Reactive Navigator.

#### *3.2.1. Reactive Module*

This module is one of the important modules for proper functioning of the wheelchair. The reactive module comprises of the following:

- Heuristic Engine,
- Memory Module and

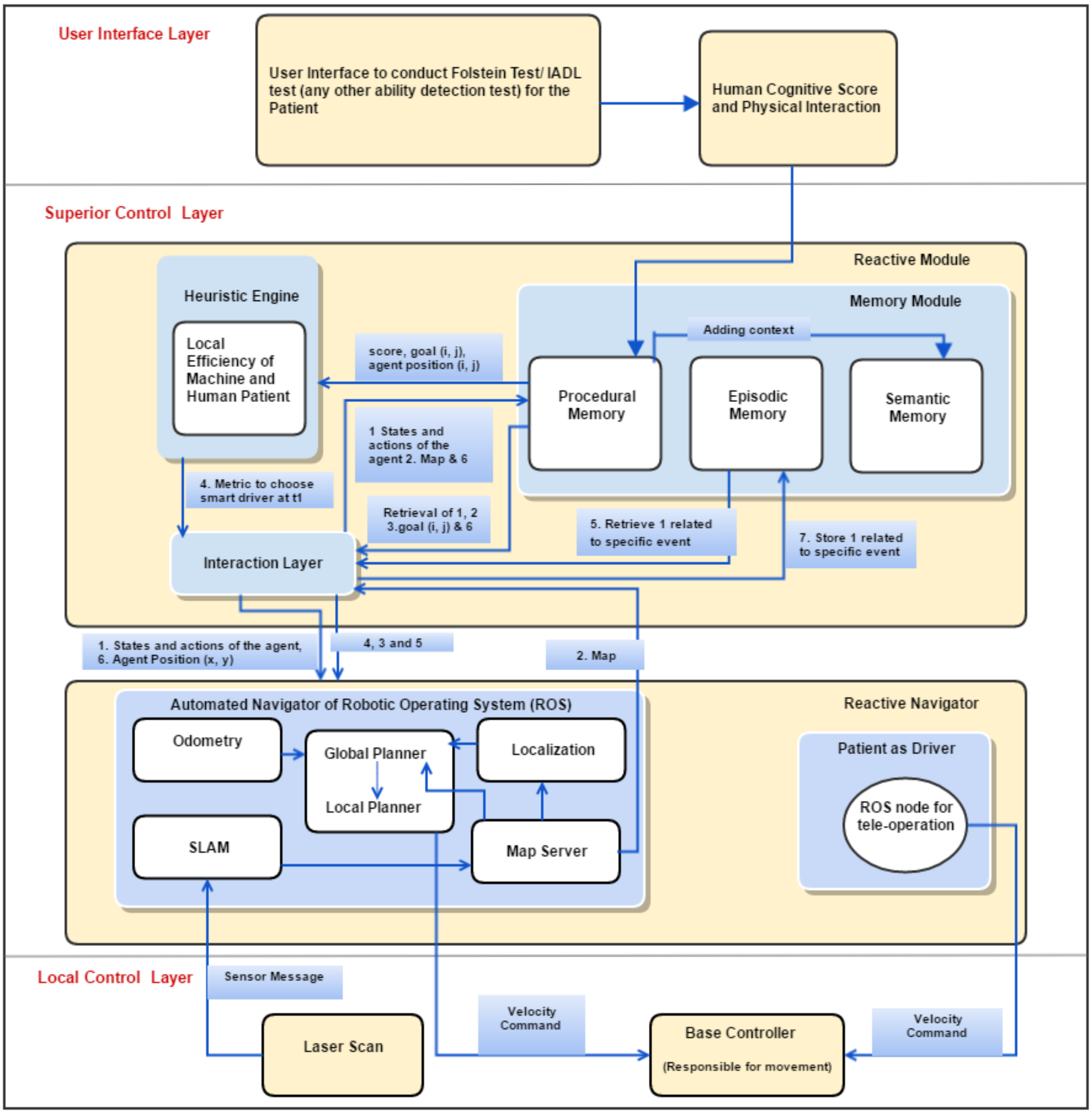

Fig. 1. Cognitive Collaborative Control Architecture

- Interaction Layer.

**1. Heuristic Engine** The control architecture given in the Figure 1 is followed by both the machine and human. These type of architectures designed for rehabilitation of patients should consist of modules which corresponds to constant monitoring of local efficiencies of human/robot command or prediction of intention of human as a driver or measuring the physiological state of the human. A method has been proposed in [9] to calculate local efficiencies of human as well as robot. Similarly in [39], an approach for intention prediction is given. Heuristic engine utilize the score from procedural memory to produce a priority configuration file. This file contains the priority of two modes of operation (human user and machine) of the P3AT robot. The users with low cognitive score should be allowed to drive the P3AT carefully. The intervention level of the system will be more and machine should take back the control from the user as and when required. On the other hand the system's intervention is less while the P3AT is driven by users with high cognitive score.

Heuristic engine uses shared goal location(i,j) and agent position(i,j) with respect to maze environment to produce a performance related metric called distance. This metric along with the priority configuration file will identify the quality of driving of the cognitive agent. The heuristic engine can guide us to model an effective collaborator.

**2. Memory Module**

- Procedural memory,
- Episodic memory and
- Semantic memory.

One important design issue related to memory module is to identify which of the outputs of different modules will go into which memory structure. Clear distinction among different memory structures is given in [13].

**a. Procedural memory**

It is concerned with how things are done [13]. In achieving a particular shared goal, it is indeed an inevitable memory structure which utilizes perception, cognitive and motor skills of the agent. State and action of agents in mobile robotics plays a vital role. Planning language help us to formalize the structure of states and actions. The states and actions of the cognitive agent will be included in procedural memory. The cognitive and physical ability score is another entry into this memory structure from the user interface layer.

**b. Episodic Memory**

It is concerned with the remembrance of personally experience event [13]. The human Wayfinding behaviour as explained in [4] follows different strategies to traverse the maze successfully. If we consider Central Point Strategy of [4], then there must be specific states and their corresponding sequence of actions that the agent tries to remember. Agent may also be interested in states and actions of two modes of operation; because in the middle of the simulation human users' incapability allows the shifting of control to automated navigator of ROS.

**c. Semantic memory**

This memory structure will provide context i.e. from procedural memory it can collect certain states and actions and based on which semantic memory try to assign meaning to the current situation.

Our future aim is to formalize the states and action of the cognitive agent. For the time being, the states are internal to ROS environment and actions of all the modules are the result of different packages concurrently executing. For successful functioning of the whole system, all packages are dependent on each other through the publisher-subscriber mechanism of ROS.

**3. Interaction Layer**
This sub-module act like a communication backbone for entire architecture. It simultaneously interacts with the heuristic engine, memory module and reactive navigator. Thus we have two communication paths.

- Heuristic Engine-Interaction Layer-Reactive Navigator
- Memory Module-Interaction Layer-Reactive Navigator

**a. Heuristic Engine-Interaction Layer-Reactive Navigator:**
Heuristic Engine provides us the smart driver at any instant of time with the help of procedural memory. From the User Interface Layer(see Figure 1), procedural memory receives cognitive score (no label), goal location (label 3 of Figure 1) and agent position(label 6 of Figure 1). These three information is utilized by heuristic engine to constitute two metrics (explained in subsection 4.1) for choosing the smart driver. When the wheelchair is run for the first time, then interaction layer access these metrics from heuristic engine to invoke anyone of the two sub-modules of the reactive navigator. The communication path is described in table 2. With the help of heuristic engine, interaction layer have to dynamically track which component of the reactive navigator has to be invoked. So, interaction layer also communicates with the memory module for states and action sequences (label 1 from Figure 1) along with the agent position(label 6 of Figure 1).

Table 2 : Detailed Specification of Heuristic Engine-Interaction Layer-Reactive Navigator Communication Path

| Input (From \| To) | Label on Architecture(figure 1) | Functionality |
|---|---|---|
| Heuristic Engine \| Interaction Layer | 4 | Metric to choose smart driver at any instant of time. |
| Reactive Navigator \| Interaction Layer | 1 | States and action of the agent |
| Reactive Navigator \| Interaction Layer | 2 | Maps of the environment generated by the `map_saver` utility of ROS. This input specifically comes from the automated navigator of ROS. Agent Position(x, y) is calculated with respect to this map. |
| Reactive Navigator \| Interaction Layer | 6 | Agent Position(x, y) is produced in the middle of simulation by the agent and stored in procedural memory via the interaction layer. |

**b. Memory Module-Interaction Layer-Reactive Navigator**
This communication path is described through table 3 and table 4. Table 3 focuses on the communication happening between interaction layer and automated navigator(mostly reactive navigator). On the other hand, table 4 focuses on the communication happening between memory module and interaction layer. Most of the labels (label 6, 4, 3, 2, 1) on the $C^3A$ (Figure 1) have already been explained in the previous communication path.

The physical interaction through the User Interface Layer(see Figure 1) leads to change in various parameters shown through different labels in the $C^3A$. Thus, it is the responsibility of the interaction layer to be aware of such kind changes in the simulation environment. Labels 7 and 5 (of Figure 1) represents the storage and retrieval of the specific states and action sequences of the agent which corresponds to specific event in episodic memory of the memory module respectively. This happens through the interaction layer and implementation related to episodic memory is kept as future work.

Table 3 : Detailed Specification of Memory Module-Interaction Layer-Reactive Navigator Communication Path

| Input (From ∣ To) | Label on Architecture(figure 1) | Functionality |
|---|---|---|
| Interaction Layer ∣ Reactive Navigator | 4 | Metric to choose smart driver at any instant of time. |
| Interaction Layer ∣ Reactive Navigator | 3 | Goal position is used by the agent to move to the shared goal |
| Interaction Layer ∣ Reactive Navigator | 5 | Specific action sequences and states of any of the agent related to a particular event. |
| Interaction Layer ∣ Reactive Navigator | 6 | Agent's localization information and is vital for the agent to trace the progress in achieving the shared goal. |
| Interaction Layer ∣ Reactive Navigator | 1 | States and action of the agent |
| Interaction Layer ∣ Automated Navigator | 2 | Map of the maze environment and used by the global planner for planning the navigation strategy. Human patient can view the 2D map while reaching to goal state. This 2D map is generated by SLAM module of Automated Navigator of ROS and resembles to 3D version of the maze environment. |

Table 4 : Detailed Specification of Memory Module-Interaction Layer Communication

| Input (From ∣ To) | Label on Architecture(figure 1) | Functionality |
|---|---|---|
| Memory Module ∣ Interaction Layer | 3 | The goal (i, j) is the co-ordinate of the goal position as given by the human patient and stored in the procedural memory. |
| Memory Module ∣ Interaction Layer | 6 | The Agent Position(x, y) is retrieved by the interaction layer from procedural memory. |
| Memory Module ∣ Interaction Layer | 1,2 | Retrieved from procedural memory by the interaction layer and finally fed to the reactive navigator. |
| Memory Module ∣ Interaction Layer | 5 | Specific states and actions corresponding to specific episode. It is retrieved by interaction layer from episodic memory. |
| Interaction Layer ∣ Memory Module | 7 | It is result of the processing of the states and actions of the agent by the interaction layer. These things are combined in such a way that they collectively represent a peculiar event during simulation. |
| Interaction Layer ∣ Memory Module | 1,2,6 | Already explained above |

### *3.2.2. Reactive Navigator*

Before building a reactive navigator, system designer needs to be very focused in the speed sensitive parts of the cognitive architecture. In, improving the speed sensitive information, ROS introduces a new concept called “nodelets" which helps us to build reactive controller. These are almost similar to conventional nodes but lots of important features are embedded into it. Reactive navigator is composed of two sub-modules which is part of ROS control framework:

- Automated Navigator of ROS and
- Patient as Driver.

**a. Automated Navigator of ROS:**
ROS is composed of number of complex tools and libraries. Incorporation of details about automated navigator will make the architecture shown in figure 1 clumsy. Our motivation is that the complexity of ROS should not dominate the understanding of our control architecture.The automated navigator of ROS is the machine in our approach.

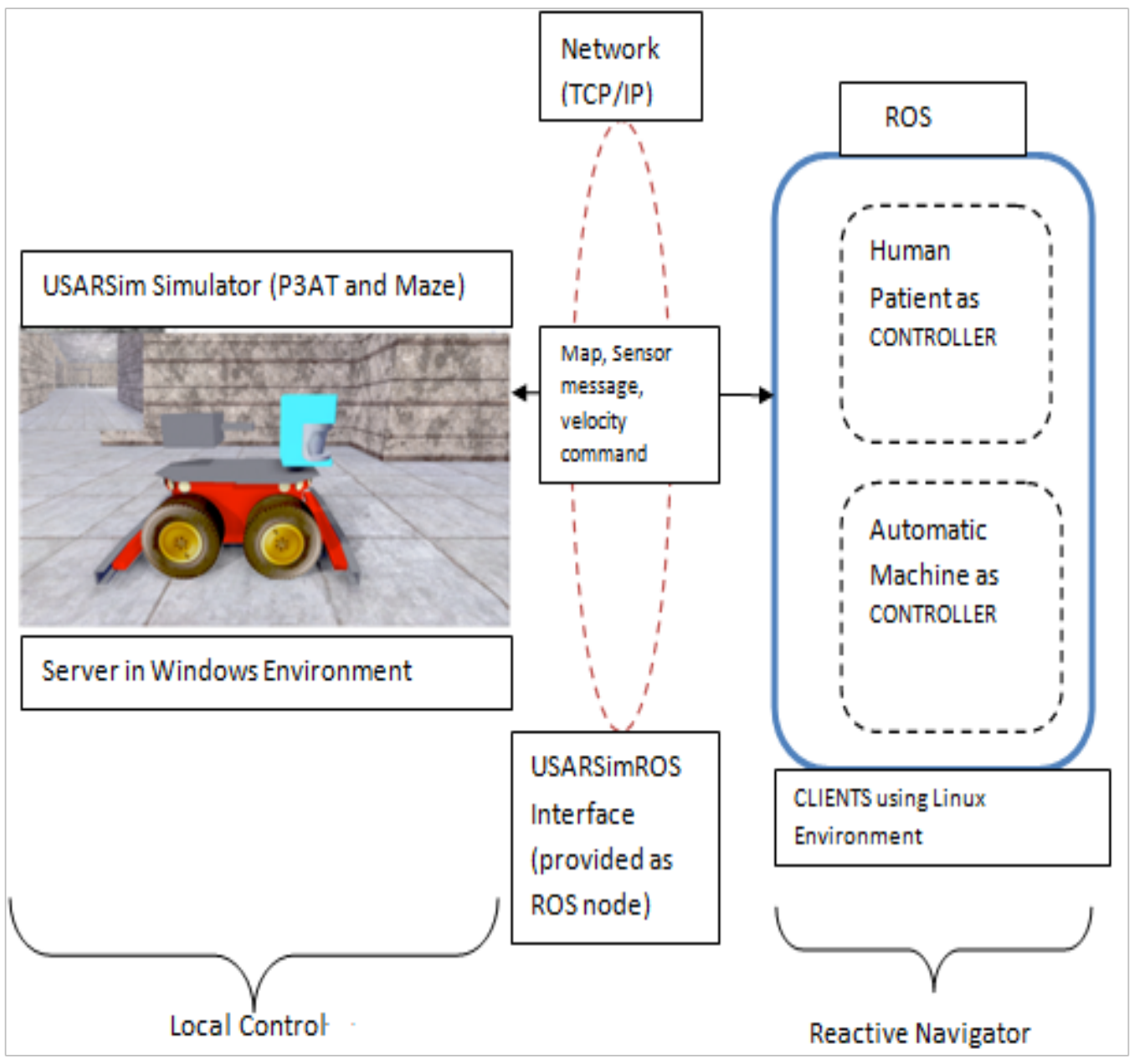

Fig. 2. Communication between USARSim Simulator and ROS

Planning is essential for guiding the agent autonomously without causing any harm. Global planner is related to creating long term plans over the entire environment [31] and it takes input from global costmap as shown in figure 3. Local planner is mainly used for obstacle avoidance which uses Local costmap. The `costmap_2d` package of ROS provides a structure which is configurable. This package maintains information about where should the robot navigate in the form of an occupancy grid [12]. Thus global planner and local planner with the information of pose estimate from AMCL (Adaptive Monte Carlo Localization) and obstacle information from costmap can plan effectively to reach the shared goal. The sensor data (in our case SICK LMS200) may be utilized in building both the costmaps.

The text that is written in communication lines (shown in figure 3) will reflect the key functionality of individual sub modules. Figure 2 describes the communication between Reactive Navigator and Local Control Layer. It depicts the communication involved in implementing our architecture integrating

USARSim and ROS control framework. It will help the reader to understand the $C^3A$ in a better way. Figure 3 describes the internal message passing between sub- modules of Automated Navigator of ROS. This figure gives detailed understanding of the how the agent can move autonomously.

The functionality of Automated navigator of ROS is achieved through a ROS node called `move_base`. This node is customized to modify the reactive behaviour (figure 3) ; so that collaboration can be achieved.

**b. Patient as Driver:**

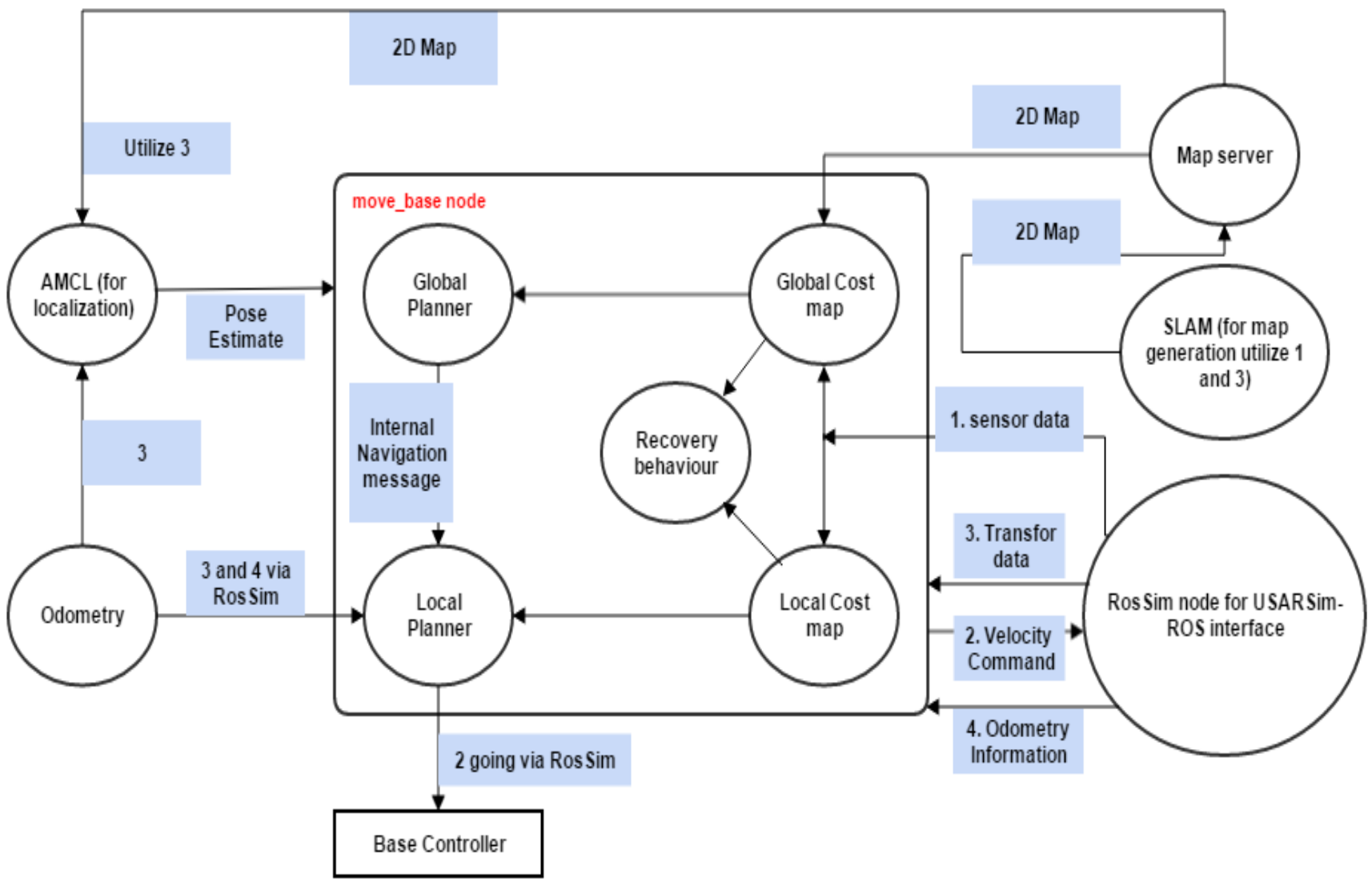

Fig. 3. Automated navigator of ROS with USARSimRos interface

This sub module uses ROS tele-operation to move around the maze environment through the keyboard. The ROS node is `teleop_twist_keyboard` [36] and corresponds to the package `brown_remotelab` [7].

### *3.3. Local Control Layer*

This layer corresponds to the 3D USARSim simulator where both the P3AT robot and maze environment co-exist simultaneously. P3AT robot uses SICK LMS200 laser range finder. It can be augmented with many other sensors. This sensor is an integral part of local control layer. All the actuators are also part of this layer. Whenever the agent is tested in real time environment, then we need a real time wheelchair. All the sensors, actuators and other hardware related aspects present in P3AT must be embedded inside the real time wheelchair.

## 4. Simulation

### *4.1. Evaluation Metrics*

There are lots of metrics [3] are suggested to test the performance of human-robot application. For experimental evaluation of the collaborative control of the P3AT robot, two types of metrics are considered.

- Condition related metric and
- Performance related metric.

Condition related metric It determines the cognitive status of the user of the wheelchair. This score leads to a priority configuration file. The ROS node, `Cognitive_Score` [see Table5] designed by us deals with the computation of this metric.
Performance related metric This metric is related to effective collaboration and finding out the efficiency of the agent while driving the P3AT. Agent drives the P3AT under two explicit modes of operation namely,

- teleoperation (human is controlling the P3AT through keyboard) and
- Machine (Automated Navigator of ROS drives the P3AT).

Implicitly collaborative mode is enforced as discussed below.

Distance between the current position of the P3AT and shared goal location with respect to the maze environment is selected as the performance metric. This distance is dynamically calculated. The `Velocity_multiplexer` node and `move_base` node [see Table5] are customized in such a way that the condition related metric along with performance related metric gives an effective way for the human to collaborate smoothly with its machine counterpart.

The system behaviour can be understood by considering two scenarios.
**Scenario: 1**
Suppose, human is controlling the P3AT and the metric indicates that P3AT is moving away from the goal. In this situation, as soon as human pauses (releases the keyboard) for a moment, then MACHINE instantly takes over the Control of P3AT. We are considering implicit communication without prior coordination. There is no explicit consensus between the two parties, but system treats this whole scenario as a situation where assistance is required. Users can always take back the control from machine whenever it is desired to do so. Thus, human is not kept away from the control loop of the system. Our aim is to allow user to utilize their own motor and cognitive skills.
**Scenario: 2**
Let us say, currently wheelchair is driven by human. Suppose the metric indicates that distance is reducing towards goal which indirectly tells about the good driving skills of the driver. So, right now even if human pauses for a moment, then MACHINE won't be able to take over the Control.

The behaviour of above two scenarios will be same when machine is considered to drive the robot. A Finite State Machine (FSM) is shown in figure 4 which will help the readers to understand the flow of the entire architecture. Figure 5 can be used to describe the two scenarios involving collaboration. Circles marked in figure 5 are the important zones. Pink cursor in Circle A shows the shared goal, the P3AT robot can be seen as the box outlined with light green boundary.
Circle A is the area where human is driving the robot well; this represent the Scenario 2 described above.

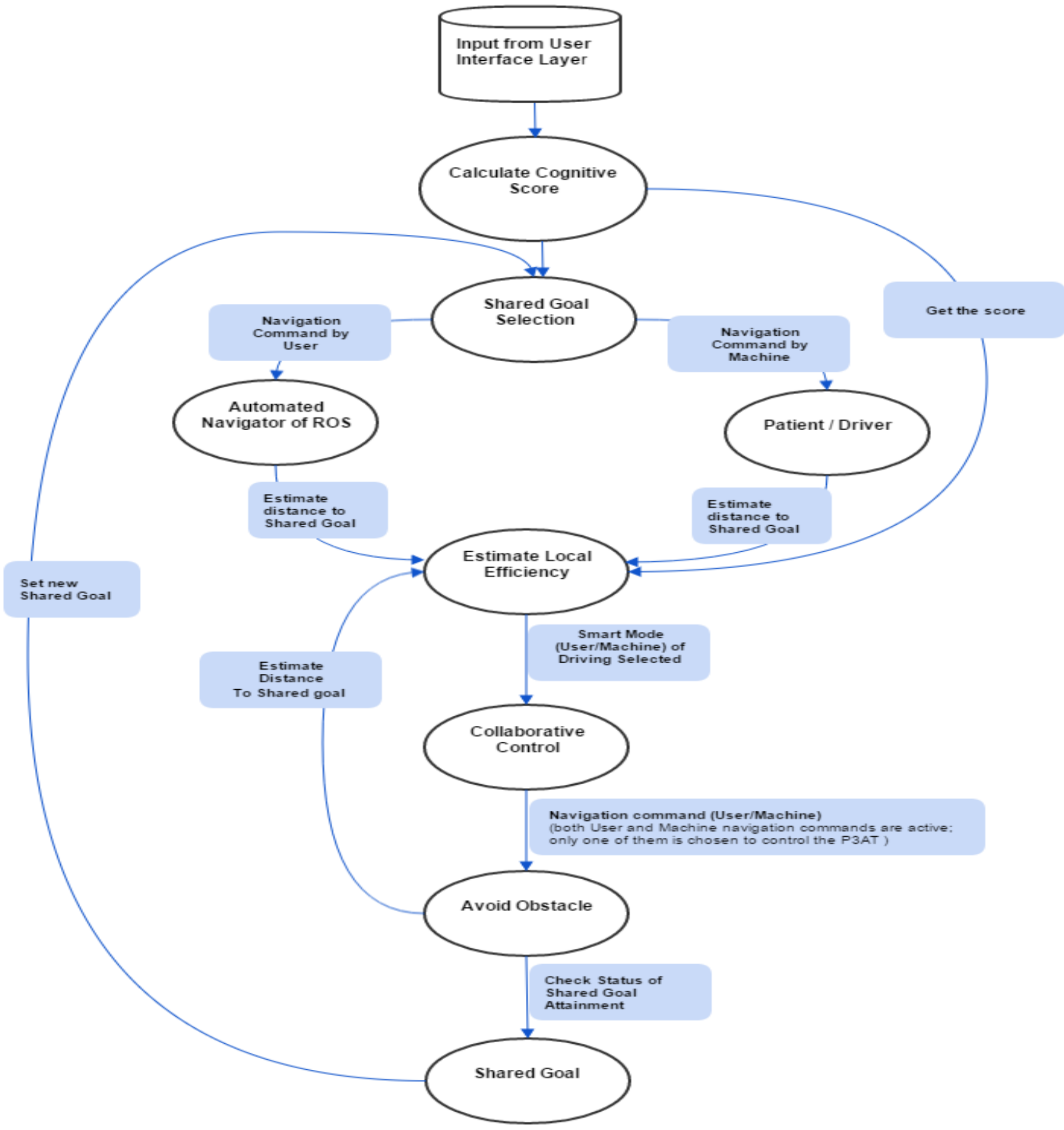

Fig. 4. FSM describing entire flow of the architecture.

Circle B and Circle C represent area where human is moving away from the shared goal. In such a situation, MACHINE takes over the control of the P3AT robot. MACHINE corrects the erratic driving behaviour of the human by taking over the control of the robot. Scenario 1 is visualized through these circles.

### *4.2. Experimental Set Up*

The experimental simulation set up is shown in figure 6. To start the services of different ROS nodes, a universal launch file is created. Simulation starts from a “form fill-up" process by the user. It is followed by launching the testing arena which is the maze environment. This maze is designed using “Unreal Development Kit" which is embedded inside USARSim simulator. The same maze is utilized for testing all the participants. Except for the testing arena, every other utility is provided through ROS nodes. Table 5 gives a detailed specification of the functionality of different ROS nodes and their correspondence with

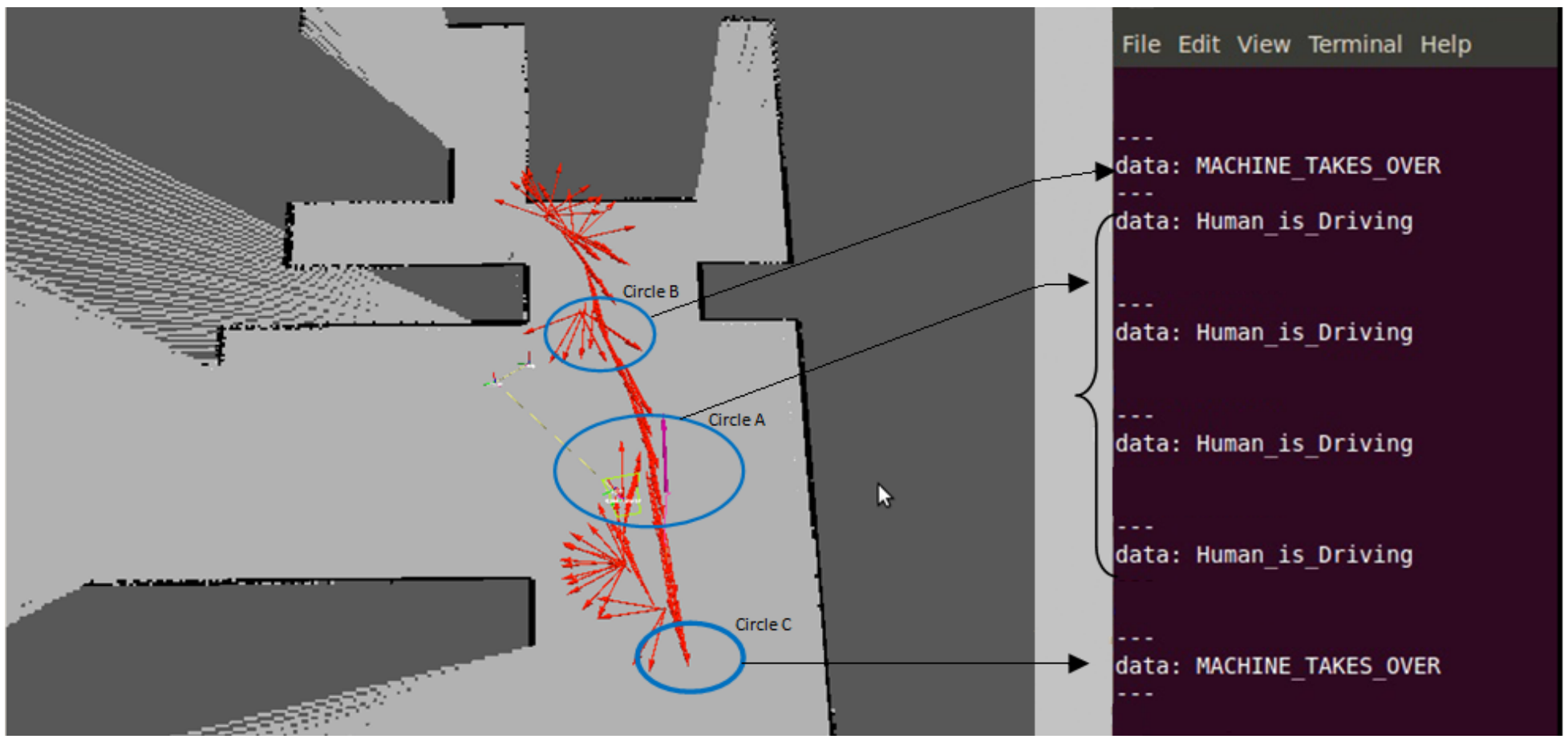

Fig. 5. Collaboration between man and machine in USARSim-ROS environment

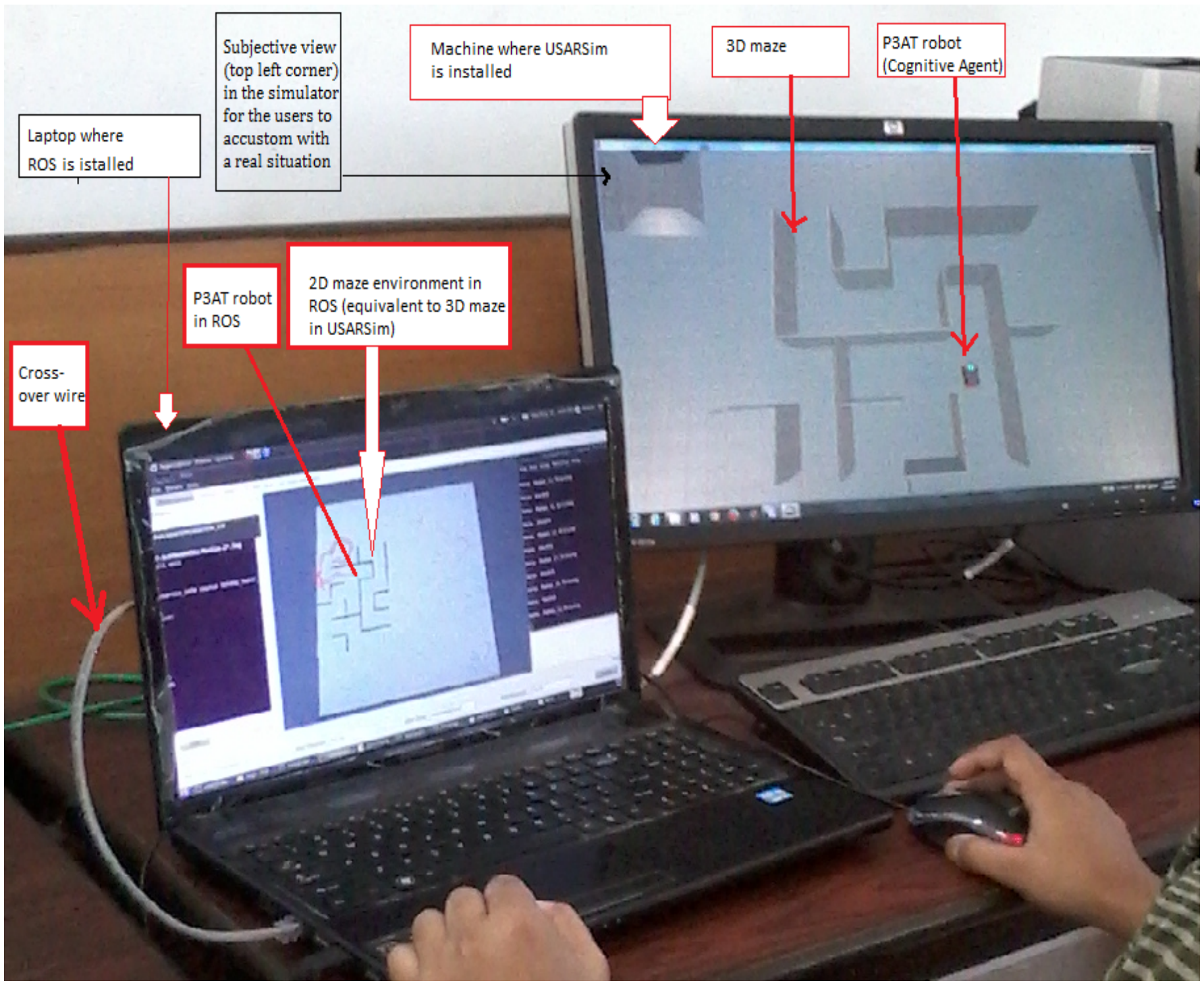

Fig. 6. Experimental simulation set up

C[3]A. Experimentation can start only after all nodes mentioned in table 5 functions properly[3].

[3]Interested readers can access https://drive.google.com/file/d/0BwgHQRnAyRbgWXhhdWxlN0ctU0k/view?usp=sharing to preview the snapshot of our whole system (the graph of the whole simulation when all ROS nodes are correctly running) in a successful scenario.

Table 5 : Detailed Specification of ROS nodes running inside the cognitive agent.

| ROS Node | Functionality | Correspondance with the Collaborative Control Architecture |
|---|---|---|
| RosSim | 1. Provides interface between ROS and USARSim, 2. helps in spawning a robot in USARSim, 3. Auto-discovering the robot's sensors, actuators | Resides in Local Control Layer and provides service to Superior Control Layer |
| teleop_twist_keyboard | With the help of it Users can drive a robot through keyboard | Local Control layer |
| amcl | amcl is a probabilistic localization system for a robot moving in 2D | Reactive Navigator |
| slam_gmapping | provides SLAM capabilities | Reactive Navigator |
| move_base | given a goal in the world, it will attempt to reach it with a mobile base. | Automated navigator of ROS |
| nodelet_manager | Implements nodelets | Reactive Module |
| Velocity_multiplexer | Multiplexes the velocity commands coming from teleop_twist_keyboard and move_base node. | Reactive Module |
| Cognitive_Score | Generate a priority configuration file which is used by Velocity_multiplexer node. | User Interface Layer |
| rviz | Visualization tool for setting shared goal | Superior Control Layer |
| map_server | allows dynamically generated maps to be saved to file. | Reactive Navigator |

The initial launching location of P3AT robot is same for all the participants. The term "shared goal" is vital for any collaborative system. We have assumed that the user can clearly locate the shared goal through the User Interface Layer. In this simulation, users can set a goal via 'rviz' window through mouse-click/touch-pad over 2D map generated through SLAM module. The "shared goal" is fixed in a corner of the maze for all participants. The 'rviz' package acts as visualisation tool for the ROS. So, in case of our $C^3$A, the goal is universal; both machine and human can rely on it.

### *4.3. Experimentation*

The cognitive collaborative control architecture is evaluated for its effectiveness by comparing the manoeuvring time (total time required from start to goal) under the following modes.
1. human(standalone mode),
2. machine (standalone mode) and
3. collaborative control mode.

The cognitive score of the subjects is based on the questionnaire from the table 6. Questions are adapted from standard papers Instrumental activities of daily living (IADL)[27] and Folstein test or mini-mental state examination (MMSE)[33]. The questionnaire can be used as an instrument for analyzing the participants in a short period of time.

Six subjects are taken in random. Each of them is made to go through the questionnaire; cognitive score is evaluated. There are two groups of people; group A having high cognitive score (HCS) and group B with low cognitive score (LCS). Four subjects belong to group A, as their cognitive score is more than five. Two persons belong to group B; as they score less than five. Table 7 shows the profile of six sub-jects that test our simulation. All the subjects are given three minutes to memorize/learn the navigation commands and setting of global shared goal. Subjects are told about their task to be completed in the simulated environment.

Table 6 Questionnaire used for cognitive score determination

| Question number | Question | Record Answer Here | Secured Score |
|---|---|---|---|
| 1 | Have you played any game before? | Yes / No | ◯Negative Response<br>◯Positive Response |
| 2 | Do you know about mazes? | Yes / No | ◯Negative Response<br>◯Positive Response |
| 3 | Can you solve a maze on paper? | Yes / No | ◯Negative Response<br>◯Positive Response |
| 4 | Can you utilize essential features of your mobile phone? | Yes / No | ◯Negative Response<br>◯Positive Response |
| 5 | Can you prepare adequate meals if supplied with ingredients? | Yes / No | ◯Negative Response<br>◯Positive Response |
| 6 | What is todayâĂŹs date? | Date | ◯Correct<br>◯Incorrect |
| 7 | What city are we in? | City | ◯Correct<br>◯Incorrect |
| 8 | Can you also tell me,what season it is? | Season | ◯Correct<br>◯Incorrect |
| 9 | What floor are we in? | Floor | ◯Correct<br>◯Incorrect |
| 10 | How many vowels in the word aeroplane? | 4/ 5 / 6 / other | ◯Correct<br>◯Incorrect |
| | Deriving the total Cognitive score,<br>i) Add 1 for each correct and positive response given by the subject,<br>ii) 0-4 = Low Cognitive Score (LCS)<br>5-10= High Cognitive Score (HCS) | | Total Score=_____<br><br>(Maximum 10) |

Table 7 : cognitive score profiling and behavioural estimation through observation

| subject_id | Cognitive Score | Group Placement | Behavior(Observed during simulation) | Cognition(Observed during simulation) | Attentiveness (Observed during simulation) |
|---|---|---|---|---|---|
| subject_1 | 10 | A | Co-operative | Good memory | Very Attentive |
| subject_2 | 3 | B | Need encouragement | Less skills related to memorizing the Commands | Attentive. |
| subject_3 | 9 | A | None | Good memory | Attentive |
| subject_4 | 10 | A | Need encouragement | Less skills related to memorizing the Commands. | Less Attentive |
| subject_5 | 2 | B | Frustrated Easily, needs encouragement | Distractible, Less skills related to memorizing the Commands | Attentive. |
| subject_6 | 10 | A | Co-operative | Good memory and expressiveness | Very Attentive |

## 5. Results

Figure 7 reports the manoeuvring time (in seconds) required by all the six subjects to reach the pre-specified goal. All the three modes described in subsection 4.3 are shown in the figure 7. Machine time for different runs is different. This is because of the way the machine navigates and users has nothing to do with this. The machine in our case is the automated navigator of ROS. This navigator generates different recovery behavior based on dynamic construction of Costmap. Costmap is built using the laser scans obtained from SICK LMS200 laser range finder. A single laser scan might not arrive in the same time window between two sets of experiment leading to a different Costmap and thus behavior. The performance of the collaborative mode versus the human user (standalone mode) is compared to evaluate the system. The autonomous navigation of ROS is highly configurable, making possible to use a large number of algorithms. The driving behavior of different users will be different if there are no traffic rules. In our maze, there is no driving rule to go to the shared goal. Thus, in two different trials, the user commands will leave the whole system in two different situations. Thus, the planners inside the move_base node acts differently which gives rise to different completion time in collaborative control mode.

As expected the time taken by the human to reach a particular goal is more as compared to driving the P3AT in collaborative mode. All the LCS group i.e. group B participants' benefits from the $C^3A$. Similar trend can be observed in most of the participants of the HCS group i.e. group A participants'. Note that an exception occurred in case of `subject_4`. Users were observed during simulation and subjective estimation done. As seen from Table 7, `subject_4` is less attentive and fails to memorize the navigation commands properly; can not relate them to arrow keys. Even though, this subject is from HCS group; its driving is erratic and frequently moves the robot away from the shared goal. Machine requires more time to make corrections to bring the robot on-course, leading to increase in manoeuvring time under collaboration.

Our architecture is not yet implemented in a real wheelchair. A differential model of the wheelchair has to be created inside USARSim to replace the P3AT robot. Several real robots works successfully by using ROS which encourage us to believe that this initial simulation result is a good starting point to build a real one. The generalizable nature of our findings is an important issue to discuss here. Our claim is not to prove the generalized nature of our simulation result on different types of wheelchair users. This initial research should be viewed as a new way to achieve the collaborative control by exploiting the open Source research platform, ROS.

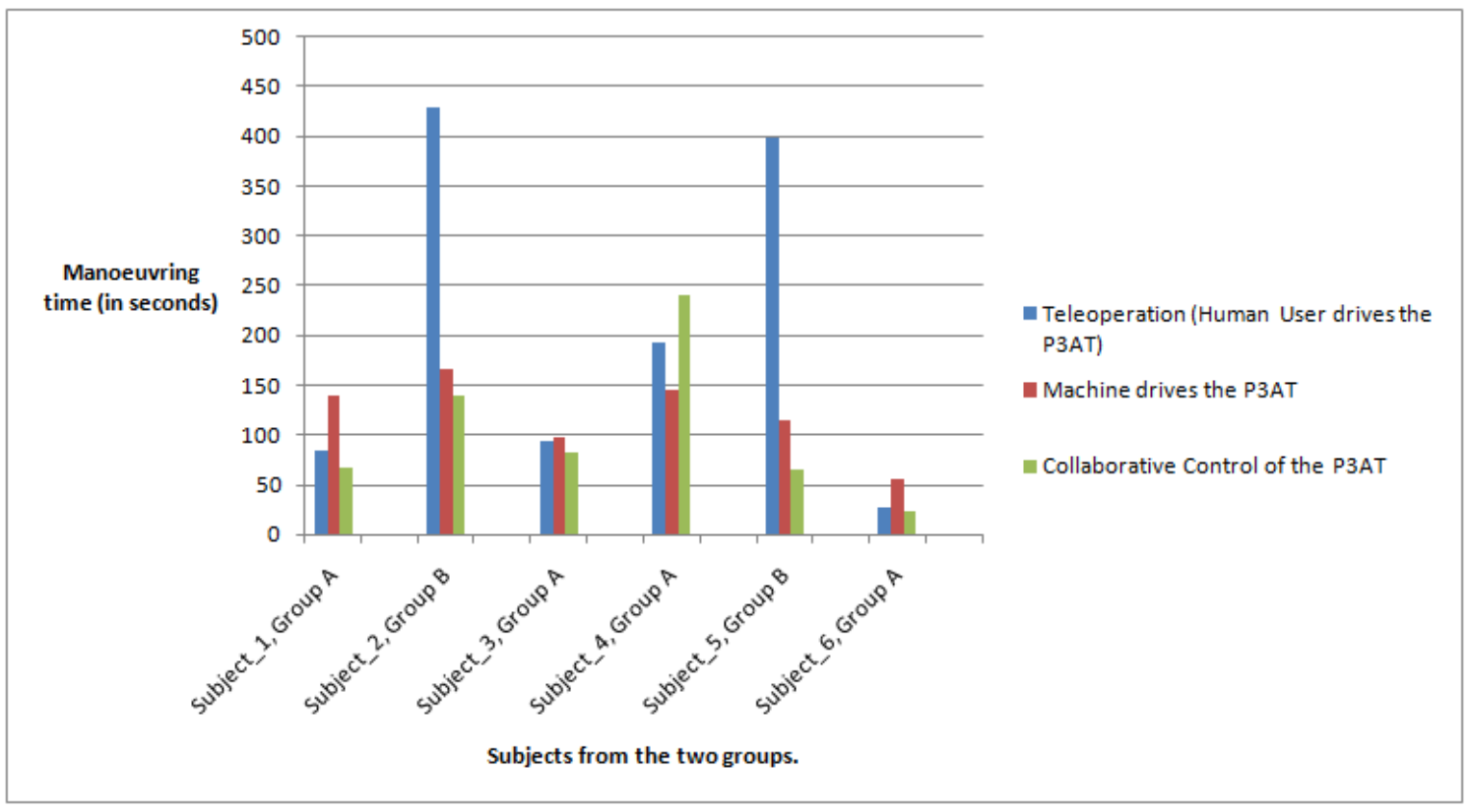

Fig. 7. Time required with different subjects under three distinct modes of operation

## 6. Final Comments

Robust real-world behavior can not be pre-programmed. ROS overcomes this limitation because of its sound message passing and shared memory concept. The wheelchair control imbibing $C^3A$ is aware of its location, driver quality along with other environmental facts such as proximity towards the shared goal through the memory module. The navigation commands from user and the wheelchair are simultaneously active through ROS topic. Assistive wheelchair with collaborative control is achieved by $C^3A$ through ROS is an approach of embodiment of cognition. ROS contains all the required features to build an effective collaborative device.

The limitation of our approach are mentioned below:

- Our initial result with six participants is used to show that our approach has some positive directions. Rigorous testing with more users are possible; once we implement the whole memory module. In our future work, we try to incorporate more evaluation metric and rigorous statistical analysis to improve our approach towards collaborative control.
- For simulation reported in this paper, the user has to be visually active to set the shared goal and have motor control to operate the keys.

The novelty of the paper is the development of $C^3A$ and the functionality of ROS is effectively utilised to analyze the utility of collaborative control. This approach for designing cognitively enabled wheelchair introduces several new research avenues within the ROS environment. Instead of using autonomous navigator of ROS, agent can be programmed to use a planner designed using formal methods. Several researchers are working on EEG (electroencephalogram) based brain-actuated wheelchair system using motor imagery. Overcoming the above mentioned limitations by extending ROS is part of our ongoing research. Controlling the wheelchair through EEG is our future goal.

## References

[1] A. Bauer, D. Wollherr and M. Buss, Human-robot collaboration: a survey, in: International Journal of Humanoid Robotics, 5(01), 2008, pp. 47-66.
[2] A. C . Schultz, Using computational cognitive models to build better human-robot interaction, in: NAE US FOE Symposium, 2006.
[3] A. Steinfeld, T. Fong, D. Kaber, M. Lewis, J. Scholtz, A. Schultz, and M. Goodrich. Common metrics for human-robot interaction, in: Proceedings of the 1st ACM SIGCHI/SIGART Conference on Human-robot Interaction (HRI '06), New York, USA, 2006, pp. 33-40.
[4] A. Saikia and S. M. Hazarika, Solving a Maze: Experimental Exploration on Wayfinding Behaviour for Cognitively Enhanced Collaborative Control, in: Intelligent Interactive Technologies and Multimedia, Springer Berlin Heidelberg, 2013, 163-177.
[5] B. D. Argall, Machine Learning for Shared Control with Assistive Machines, in: Proceedings of ICRA Workshop on Autonomous Learning: From Machine Learning to Learning in Real world Autonomous Systems, Karlsruhe, Germany, May 2013.
[6] B. Sathish Babu, (n.d.). Cognitive agents, 1st ed. [ebook] Retrieved from: pet.ece.iisc.ernet.in/sathish/cognitive.pdf
[7] C. Crick, T. Jay and S. Osentoski(2011), `brown_remotelab` - ROS Wiki. [online] Wiki.ros.org. Retrieved from: http://wiki.ros.org/brown_remotelab.
[8] C. Urdiales, From shared control to collaborative navigation, in: Collaborative Assistive Robot for Mobility Enhancement (CARMEN), volume 27 of Intelligent Systems Reference Library, Springer Berlin Heidelberg, 2012, pages 41-66.
[9] C. Urdiales, et al., Wheelchair collaborative control for disabled users navigating indoors, in: Artif. Intell. Med, Vol. 52, 2011, pp. 177-191.
[10] D. Kortenkamp. Designing an architecture for adjustably autonomous robot teams, in: R. Kowalczyk, S. Loke, N. Reed, and G. Williams, editors, Advances in Artificial Intelligence, PRICAI 2000 Workshop Reader, volume 2112 of Lecture Notes in Computer Science, Springer Berlin Heidelberg, 2001, 335-338.
[11] E. Marder-Eppstein, (2012). `move_base` - ROS Wiki. [online] Wiki.ros.org. Retrieved from: http://wiki.ros.org/move_base.
[12] E. Marder Eppstein, (2010). `costmap_2d` - ROS Wiki. [online] Wiki.ros.org. Retrieved from: http://wiki.ros.org/costmap_2d.
[13] E. Tulving, Memory and consciousness, in: Canadian Psychology/Psychologie canadienne, Vol 26(1), Jan 1985, 1-12, doi:10.1037/h0080017.
[14] F. Heger, L. Hiatt, B. Sellner, R. Simmons, and S. Singh, Results in sliding autonomy for multi-robot spatial assembly, in: 8th International Symposium on Artificial Intelligence, Robotics and Automation in Space (iSAIRAS), 2005, 5-8.
[15] G. E. Gresham et al., Prevention and rehabilitation of stroke, in: Stroke, 28(7), 1997, pp. 1522-1526.
[16] J. A. Adams, P. Rani, and N. Sarkar (2004). Mixed initiative interaction and robotic systems, in: Workshop on Supervisory Control of Learning and Adaptive Systems, in: Nineteenth National Conference on Artificial Intelligence (AAAI-04), San Jose, CA, USA.
[17] J. Brookshire, S. Singh and R. Simmons, Preliminary results in sliding autonomy for coordinated teams, in: Proceedings of The 2004 Spring Symposium Series, 2004.
[18] J. E. Allen and C. Guinn, Mixed-initiative interaction, in: Intelligent Systems and their Applications, IEEE, 14(5), 1999, 14-23.
[19] J. R. Anderson and C. Lebiere, "The Newell test for a theory of cognition." in: Behavioral and brain Sciences, 26 (05), 2003, pp. 587-601.
[20] J. Laird, A. Newell and P. Rosenbloom, Soar: An architecture for general intelligence, in: Artificial Intelligence, 33 , 1987, 1-64.
[21] J. W. Kim and F. .E. Ritter (2012). ACT-R Frequently Asked Questions List. [online] Acs.ist.psu.edu. Retrieved from: http://acs.ist.psu.edu/projects/act-r-faq/act-r-faq.html
[22] K. Kawamura, P. Nilas, K. Muguruma, J. Adams, and C. Zhou, An agent-based architecture for an adaptive human-robot interface, in: Proceedings of the 36th Annual Hawaii International Conference on System Sciences, 2003, pp. 126.2.
[23] L. Fehr, W. E. Langbein, and S. B. Skaar, Adequacy of power wheelchair control interfaces for persons with severe disabilities : A clinical survey, in: Journal of Rehabilitation Research and Development, 37(3), 2000, pp. 353-360.
[24] M. A. Goodrich and A. C. Schultz, Human-robot interaction: a survey, in: Foundations and trends in human-computer interaction, 1(3), 2007, pp. 203-275.
[25] M. Johnson, J. M. Bradshaw, P. J. Feltovich, C. M. Jonker, B. van Riemsdijk, and M. Sierhuis, The fundamental principle of coactive design: interdependence must shape autonomy, in: Coordination, Organizations, Institutions, and Norms in Agent Systems VI, Springer Berlin Heidelberg, 2011, 172-191.
[26] M. N. Huhns and M.P. Singh, Cognitive agents, IEEE, in: Internet Comput., November/December ( 1998), pp. 87-89.

[27] M. Lawton and E. Brody, Assessment of older people: self-maintaining and instrumental activities of daily living, in: Gerontologist, 9, 1969, 179-85.
[28] M. Quigley, et al., ROS: an open-source Robot Operating System, in: ICRA workshop on open source software, Volume. 3. No. 2, 2009.
[29] M. Wooldridge and N. R. Jennings, Agent Theories, Architectures, and Languages: a Survey, in Wooldridge and Jennings eds., Intelligent Agents, Springer-Verlag, 1995, 1-22.
[30] P. Langley, An adaptive architecture for physical agents, in: Proc. IEEE/WIC/ACM Int. Conf. Intell. Agent Technol., Compiegne, France, 2005, pp. 18-25.
[31] P. Yoonseok(2014), Navigation/Tutorials/Robotsetup-ROS Wiki, [online] Wiki.ros.org. Retrieved from: http://wiki.ros.org/navigation/Tutorials/RobotSetup.
[32] R. A. Brooks., A robust layered control system for a mobile robot, in: IEEE Journal of Robotics and Automation, 2(1), 1986, pp. 14-23.
[33] R. Crum , J. Anthony, S. Bassett and M. Folstein, "Population-based norms for the mini-mental state examination by age and educational level.", in: Journal of the American Medical Association, 269(18), 1993, pp. 2386-3239.
[34] R. Simpson, Smart wheelchairs: A literature review., in: J. Rehabil. Res. Dev., 42(4), 2005, 423-436.
[35] S. Balakirsky (2013). *USARSim.* [online] SourceForge. Retrieved from: http://sourceforge.net/projects/usarsim/.
[36] S. B. Balakirsky and Z. Kootbally, USARSim/ROS: A Combined Framework for Robotic Control and Simulation, in: Proceedings of the ASME 2012 International Symposium on Flexible Automation (ISFA 2012), St. Louis, June 18-20, 2012.
[37] S. J. Russell and P. Norvig, Artificial Intelligence, Englewood Cliffs, N.J.: Prentice Hall, 1995. Print.
[38] S. Rosenthal, and M. Veloso, Using symbiotic relationships with humans to help robots overcome limitations, in: Workshop for Collaborative Human/AI Control for Interactive Experiences, 2010.
[39] T. E. Carlson, Collaborative control mechanisms for an intelligent robotic wheelchair, Ph.D. Dissertation, Imperial College London (University of London), 2010.
[40] B. D. Argall, Modular and adaptive wheelchair automation, in: Experimental Robotics, Springer International Publishing, 2016.
[41] Y. Kang, et al., Stability analysis of a class of hybrid stochastic retarded systems under asynchronous switching, in: IEEE Transactions on Automatic Control, 59 (6), 2014, pp. 1511-1523.
[42] Y. Kang, et al., On Input-to-State Stability of Switched Stochastic Nonlinear Systems Under Extended Asynchronous Switching, in: IEEE transactions on cybernetics, 46 (5), 2016, pp. 1092-1105.
[43] R. Zhang et al., Control of a Wheelchair in an Indoor Environment Based on a BrainâĂŞComputer Interface and Automated Navigation, in: IEEE transactions on neural systems and rehabilitation engineering, 24 (1), 2016, pp. 128-139.
[44] K. Ericson, S. Pallickara, and C. W. Anderson, Analyzing electroencephalograms using cloud computing techniques, in: Proceedings of the Second (IEEE) International Conference on Cloud Computing Technology and Science (CloudCom), USA, 2010.